\documentclass[runningheads]{llncs}
\usepackage[T1]{fontenc}
\usepackage{graphicx,verbatim}

\usepackage{times}
\usepackage{epsfig}
\usepackage{amsmath}
\usepackage{amssymb}
\usepackage{booktabs}
\usepackage{multirow}
\usepackage{ulem}
\usepackage{xcolor}
\usepackage[hidelinks]{hyperref}

\newcommand{\subpara}[1]{\vspace{0.4em} \noindent \textbf{#1}}

\begin{document}
\title{Registration is a Representation Problem: Solving Whole-Body PET/CT Registration}
\titlerunning{Registration is a Representation Problem}

\author{Neel Dey \and Adrian V. Dalca}
\authorrunning{Dey and Dalca}
\institute{
A. A. Martinos Center for Biomedical Imaging, Massachusetts General Hospital \\
Department of Radiology, Harvard Medical School \\
Computer Science and AI Lab, Massachusetts Institute of Technology 
    \email{\{ndey,adalca\}@mgh.harvard.edu}}
  
\maketitle
\begin{abstract}
Raw image intensities are a poor surrogate for similarity across images in correspondence problems. Further, while auxiliary segmentation labels can provide semantic identification and improve matching, they can assign only coarse identities to fine-grained anatomy and often cover only a subset of structures, particularly in whole-body settings. We present an approach that instead establishes correspondence using complementary representations of anatomy and demonstrate its utility for longitudinal whole-body PET/CT registration in the Learn2Reg 2026 PSMAReg Challenge. 
We use our 3D foundation model, anatomix, to derive fine-grained, per-voxel semantic representations that are consistent across volumes, and further employ off-the-shelf segmentation models to obtain coarse anatomical identities. Together, these representations drive a GPU-accelerated iterative registration solver (FireANTs) in a hybrid multistage framework: the first stage estimates alignment from dense multi-channel features, after which a second refines the deformation using both anatomix- and segmentation-based losses. 
Lastly, because registration should not introduce non-biological tumor volume changes across time, we develop lesion-specific, component-wise volume preservation schemes.
Without \textit{any} task-specific training, our system placed first on the PSMAReg Challenge leaderboard. Code and interactive visualizations are available at \url{https://www.neeldey.com/reps-for-reg}.


\end{abstract}

\section{Introduction}

The medical image registration literature's focus has largely shifted from optimization methods~\cite{avants2008symmetric} toward learned models that directly predict deformation fields~\cite{balakrishnan2019vxm} quickly and accurately within domain. When trained on diverse data, these models can further act as generalist networks that operate across imaging contexts~\cite{hoffmann2021synthmorph,tian2024unigradicon}. However, such generalists still underperform task-specific networks, necessitating training new models for every dataset, modality, or application for optimal performance. Fast registration solvers~\cite{jena2026adaptive,jena2025scalable,siebert2024convexadam} are attractive in this regard because their transformation models, objectives, and regularizers can be adapted directly to the problem at hand without requiring task-specific training~\cite{jena2024deep}. 
We show that these solvers are not intrinsically limited by their optimization methods, but by the representation used to establish correspondence.

We therefore approach registration as a representation problem. Instead of relying solely on image intensities or labels, we show that better registration can be achieved by matching features extracted from a generalist foundation model. We use anatomix~\cite{dey2025learning}, a 3D biomedical foundation model pretrained on abundant, highly variable synthetic data to extract semantically meaningful voxelwise representations across imaging modalities, anatomical regions, and downstream tasks. These features are approximately invariant to nuisance appearance variation while remaining equivariant to spatial transformations, making them naturally suited for establishing correspondence between volumes. Further, rather than relying on any single similarity signal, we combine complementary, domain-specific representations to improve longitudinal PSMA PET/CT registration for the PSMAReg challenge. We augment anatomix with MIND-SSC~\cite{heinrich2013towards}, a handcrafted descriptor based on local self-similarity, alongside anatomical and lesion segmentations that provide coarse semantic identities for individual structures. For PSMAReg specifically, we further develop a region-based matching objective for poorly overlapping organs and lesion-aware volume-preserving constraints to prevent registration from altering quantitative PET biomarkers. This combination retains the flexibility of task-specific optimization with high-quality registration solvers (like FireANTs~\cite{jena2026adaptive}) while obtaining its correspondence signal from general-purpose pretrained representations, all without any training or finetuning on the challenge's training dataset.  Our code is available at \url{https://github.com/neel-dey/reps-for-reg/}.

\begin{figure}[!t]
    \centering
    \includegraphics[width=0.98\linewidth]{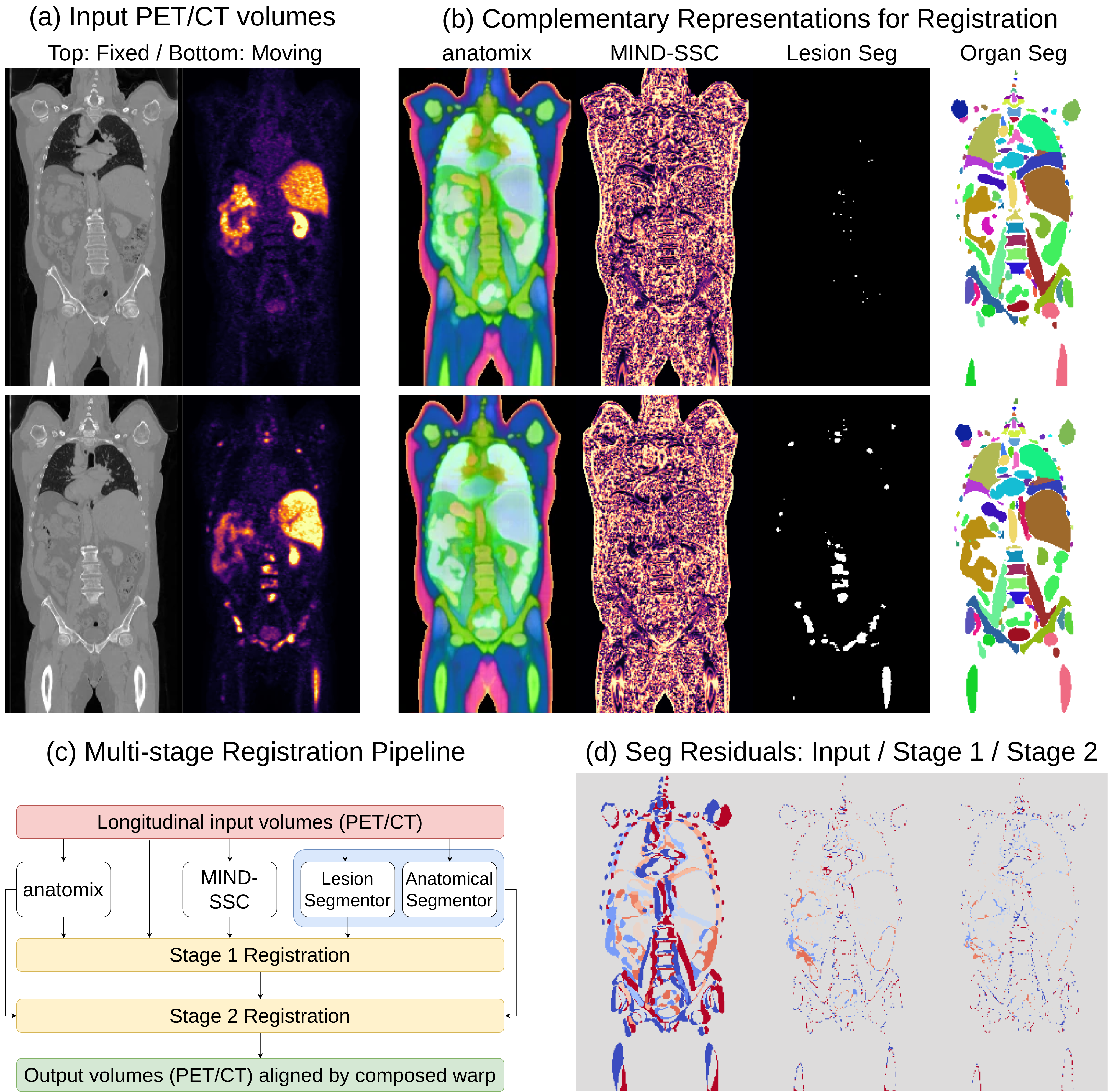}
    \caption{Given a pair of PET/CT volumes to register (\textbf{a}), we extract both fine-grained representations with the learned anatomix foundation model~\cite{dey2025learning} and the hand-crafted MIND-SSC descriptor~\cite{heinrich2013towards} and coarse-grained representations as in PSMA lesion and anatomical segmentation maps, obtained using LION~\cite{LION} and TotalSegmentator~\cite{wasserthal2023totalsegmentator}, respectively (\textbf{b}). Using these complementary representations, we design a hybrid multi-stage registration framework that uses a fast registration solver (FireANTs) to align the representations (\textbf{c}). Most residual misalignment after stage-1 is in small abdominal organs and is corrected by stage-2 (\textbf{d}). }
    \label{fig:pipeline}
    \vspace{-3em}
\end{figure}

\section{Methods}

Our two-stage pipeline is visualized in Figure~\ref{fig:pipeline}.
Stage 1 uses the images and their nonlinear representations. Stage 2 further uses segmentation losses for the optimization. Each stage is followed by a PET biomarker preserving post-processing step. 

\subpara{Feature engineering.}
\looseness=-1
We extract a set of complementary representations from the input CT volumes that capture anatomy at different levels of granularity (Fig.~\ref{fig:pipeline}b). We first use anatomix~\cite{dey2025learning} (with the publicly available \texttt{anatomix-dev-vit} weights) to extract a nonlinear 32-dimensional feature vector at each voxel. For visualization in Fig.~\ref{fig:pipeline}b, we project these features onto their top three principal components. For better performance, we complement these learned features with MIND-SSC~\cite{heinrich2013towards}, a hand-crafted descriptor based on local self-similarity that is likewise designed to be robust to cross-modal appearance variation. Finally, we obtain and use coarse semantic representations from pretrained segmentation networks: TotalSegmentator~\cite{wasserthal2023totalsegmentator} provides anatomical labels from CT, while LION~\cite{LION} provides PSMA lesion predictions from PET. Together, these learned, hand-crafted, and segmentation-derived representations provide complementary fine- and coarse-grained anatomical information for the registration stages below.

\subpara{Multi-stage iterative optimization.}
Given the representations defined above for both moving and fixed volumes, we take a hybrid approach and use an off-the-shelf GPU-accelerated registration solver (FireANTs~\cite{jena2026adaptive,jena2025scalable}) to optimize a deformation field. We take a two-stage registration approach.  \uline{Stage 1} performs feature-based alignment to remove large deformations between volumes. It stacks the CT and PET volumes alongside the aforementioned anatomix and MIND-SSC features into a multi-channel feature volume for both fixed and moving volumes ($F_f$ and $F_m$, respectively) and performs greedy deformable registration by optimizing
$\phi_1^{*}=\arg\min_{\phi_1}\;\mathcal{L}_{\mathrm{LNCC}}\!\left(F_f,F_m\circ\phi_1\right)+\lambda_1 \mathcal{L}_{\mathrm{vol}}(\phi_1,M)$, where $\mathcal{L}_{\mathrm{LNCC}}$ is the body-masked squared local normalized cross correlation matching loss~\cite{avants2011reproducible} applied independently to each channel, $\phi_1$ is the stage-1 warp, $\mathcal{L}_{\mathrm{vol}}$ is a lesion-volume preserving regularizer on $\phi$ defined on the moving image's lesion mask $M$ (described further below), and $\lambda_1 = 10$. We optimize this objective with FireANTs and regularize the solution by Gaussian smoothing of the loss gradient and deformation field with standard deviations of 2.0 and 0.5 voxels, respectively, with a step size of 2.0 and a resolution pyramid of $6\times4\times2\times1$.

While the outputs of Stage 1 are well aligned visually, we find that there is substantial headroom left in organ alignment, as small abdominal organs can be mistaken for each other when using highly shape-biased feature extractors. Therefore, \uline{Stage 2} performs segmentation-guided refinement to improve the alignment of individual organs and lesions after the large deformations have been removed. We first warp the moving organ segmentations with $\phi_1^{*}$ and represent each structure in the fixed and moving volumes using a distance transform truncated at $20$~mm and a soft boundary shell with a width of $2.0$ voxels. We also add the LION lesion segmentation as a separate structure for both moving and fixed volumes, with a weight of $0.25$. We then optimize a residual deformation $\phi_2$ using FireANTs on the following objective:
  \begin{align*}
  &\phi_2^{*}
  =\arg\min_{\phi_2}\;
  0.60\,\widehat{\mathcal{L}}_{\mathrm{GenDice}}\!
  \left(S_f,S_m\circ\phi_1^{*}\circ\phi_2\right)
  +0.15\,\widehat{\mathcal{L}}_{\mathrm{cham}}\!
  \left(S_f,S_m\circ\phi_1^{*}\circ\phi_2\right) \\
  &+0.25\,\widehat{\mathcal{L}}_{\mathrm{region}}\!
  \left(S_f,S_m,\phi_1^{*}\circ\phi_2\right) +0.5\,\widehat{\mathcal{L}}_{\mathrm{LNCC}}\!
  \left(A_f,A_m\circ\phi_1^{*}\circ\phi_2\right)
  +10\,\mathcal{L}_{\mathrm{vol}}\!
  \left(\phi_2,M\circ\phi_1^{*}\right).
  \end{align*}
Here, $S_m$ and $S_f$ are the moving and fixed multichannel segmentation maps, $A_f$ and $A_m$ are the anatomix features, and each hatted loss is normalized by its value before stage-2 optimization. The generalized Dice loss~\cite{sudre2017generalised} $\mathcal{L}_{\mathrm{GenDice}}$ does label matching while accounting for imbalance in size between large and small organs, while the one-sided Chamfer loss $\mathcal{L}_{\mathrm{cham}}$ is the mean distance (capped at 20 mm), from each warped moving structure (with its soft shell) to the fixed structure. We also add a region-based loss that measures whether tissue inside each fixed organ maps into the corresponding moving organ. For each organ, we sample up to 4,000 fixed-organ points and map them through the current deformation; each mapped point has zero error if it lies inside the moving organ and is otherwise penalized by its distance to the organ in millimeters (again, capped at 20 mm). This distance provides a useful direction for correction when the organs have little overlap and Dice gives a weak gradient. Lastly, the anatomix term stabilizes regions that are not covered by the organ labels, and the volume term limits additional lesion-volume changes during refinement.

We optimize $\phi_2$ with the compositive deformation model at one resolution level, using at most $35$ iterations, a step size of $2.0$, and Gaussian gradient and deformation smoothing with standard deviations of $1.0$ and $0.5$ voxels, respectively, before composing the final deformation as $\phi^{*}=\phi_1^{*}\circ\phi_2^{*}$ and postprocessing it as described below.

\begin{figure}[t]
    \centering
    \includegraphics[width=1\linewidth]{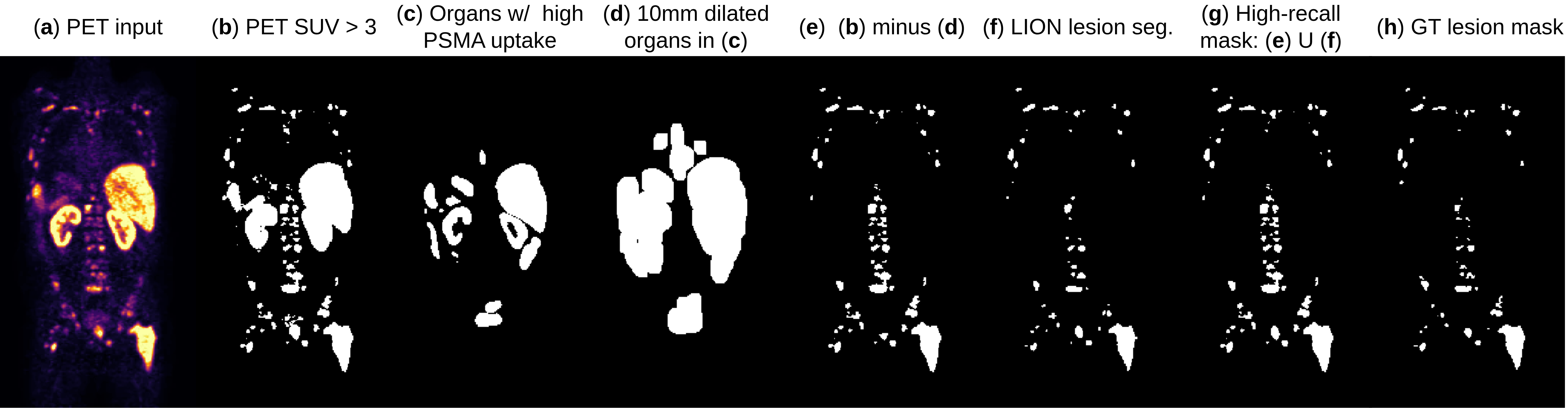}
    \caption{\textbf{Constructing a recall-biased lesion mask}. False negative PET lesion predictions would reduce PET biomarker preservation in our work and we therefore bias towards recall. Given the PET scan (\textbf{a}), we threshold it to find areas with high uptake (\textbf{b}) and subtract organs with physiological uptake (\textbf{c} and \textbf{d}) from this mask to arrive at a morphologically constructed mask (\textbf{e}). Finally, we add the segmentation prediction of a pretrained PSMA lesion segmentor (\textbf{f}) to the morphologically constructed mask to obtain the final recall-biased lesion mask (\textbf{g}).}
    \label{fig:high-recall-lesion}
    \vspace{-2em}
\end{figure}

\subpara{Volume preservation and PET biomarkers.}
Tumor burden may change biologically between the moving (follow-up) and fixed (baseline) scans, but the registration itself should not introduce additional changes in lesion volume. This is particularly important for PSMAReg, where the MTV (Metabolic Tumor Volume) and TLG (Total Lesion Glycolysis) biomarkers attain their ideal values when the deformation preserves the volume of ground-truth lesions. We therefore explicitly constrain local volume change within regions likely to contain tumor. Because these constraints can only preserve lesions included in their support, false-negative lesion predictions are particularly harmful to MTV and TLG, motivating a recall-biased lesion mask in the moving image.

While existing PSMA lesion segmentors such as LION~\cite{Piresjnumed.126.272852,LION} provide good lesion specificity, they may also miss lesions. To improve recall, we first threshold PET at SUV~$>3$ (a threshold used in some PSMA segmentors~\cite{gafita2019qpsma}). This thresholding also captures organs with high physiological tracer uptake, so we subtract masks for 16 such organs (see the Appendix) predicted from CT
using TotalSegmentator, after dilating their union by 10~mm to account for within-timepoint PET/CT misalignment. Because SUV thresholding may still miss lesions with weaker uptake, we finally take the union of this region-constrained SUV mask with the moving-time-point LION prediction. The resulting mask $M$ therefore combines organ-filtered SUV thresholding with trained lesion segmentations to improve recall and is visualized in Figure~\ref{fig:high-recall-lesion}.

Given the lesion mask $M$, we encourage volume preservation during both registration stages using $\mathcal{L}_{\mathrm{vol}}=\left\|M\ast\left(\det(\mathrm{Jac}(\phi))-1\right)\right\|_2^2/\|M\|_1$ as a regularizer~\cite{rohlfing2003volume,staring2007rigidity},
which penalizes local volume change under the deformation $\phi$. To further drive down volume change after registration, we apply the volume-preserving post-processing of~\cite{hameeteman2008volume}, which drives the deformation to locally preserve volume via a local correction applied to $\phi$, obtained by solving a Poisson system with FFTs in $<0.5$~s on a GPU. However, applying this correction to all of $M$ can reduce anatomical alignment without improving biomarker preservation for regions of $M$ where volume is already preserved. We therefore modify~\cite{hameeteman2008volume} to post-process $M$ selectively, only when a lesion component is changing its volume above a threshold. For each connected component $C_i$ of $M$, let $B_i=|C_i|$ be its original voxel count and $A_i=|C_i\circ\phi|$ its count after warping. Components under the $\left|\log((A_i+1)/(B_i+1))\right|\leq0.03$ threshold are considered adequately preserved by the warp and are not postprocessed. The rest receive up to four local volume correction passes, followed by eight component-wise passes that scale each lesion's volume until its warped voxel count $A_i$ approaches its original count $B_i$.

\subpara{Background fold correction.}
Because the image-matching losses are evaluated only within the patient, the deformation is weakly constrained in the surrounding air and can form anatomically meaningless folds. After Stage 2, on training pairs above the challenge's percentage threshold of non-diffeomorphic volume (\%NDV)~\cite{liu2024finite}, 99.5 to 100\
of folded voxels lie outside the body mask. We therefore repair folds (similar to~\cite{karaccali2004estimating}) in the background  \textit{outside} the fixed-image body mask dilated with a $5$-mm safety margin. We do six passes of expanding the detected region by two voxels and replacing the deformation there with a feathered blend of the original and Gaussian-smoothed fields ($\sigma=2.0$ voxels); the blending weight is zero within the protected body region. 

\begin{figure}[t]
    \centering
    \includegraphics[width=0.98\linewidth]{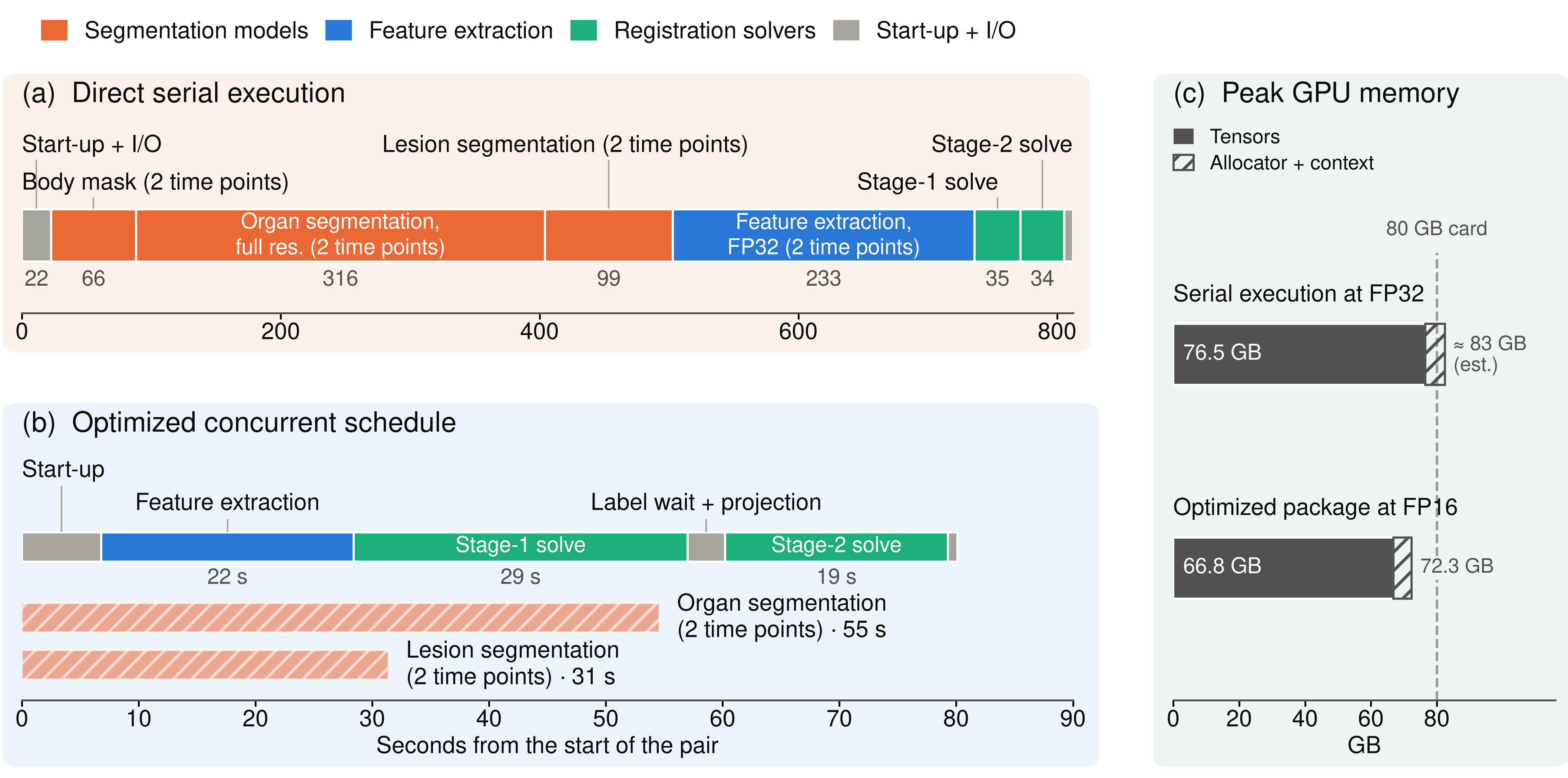}
    \caption{\textbf{Parallelizing feature extraction and segmentation network passes}. \textbf{(a)} Serially running all pretrained networks used in our framework at their default hyperparameters would exceed the 90~s limit. \textbf{(b)} We use an optimized schedule and vary the inference hyperparameters of each network to run networks in parallel and deliver segmentations before they are needed by Stage 2. \textbf{(c)} FP16 feature extraction and accumulation, plus the removal of a redundant segmentor, bring peak vRAM under the 80GB maximum defined by the challenge.}
    \label{fig:efficiency}
    \vspace{-1.5em}
\end{figure}

\subpara{Maximizing throughput and utilization.} 
The PSMAReg test phase requires each image pair to be registered within $\sim90$ s on an 80 GB H100 GPU. Because our pipeline combines multiple feature extraction and segmentation models alongside registration solvers, a direct serial implementation would vastly exceed these constraints. We therefore optimize memory use, parallelize independent components, and selectively trade model resolution for speed, reducing the full pipeline from 83 GB and 812~s per pair to 72 GB and 79~s. In particular, we restrict computation to relevant image regions, run feature extraction and segmentation concurrently, use FP16 precision and gradient checkpointing where possible, remove overhead from library startup, file compression, and repeated CPU-to-GPU transfers and use faster, lower-quality segmentation models (e.g., TotalSegmentator's 6-mm model) in stages where high-resolution predictions are unnecessary. The resulting execution schedule is shown in Figure~\ref{fig:efficiency}.

\begin{figure}[!t]
    \centering
    \includegraphics[width=\linewidth]{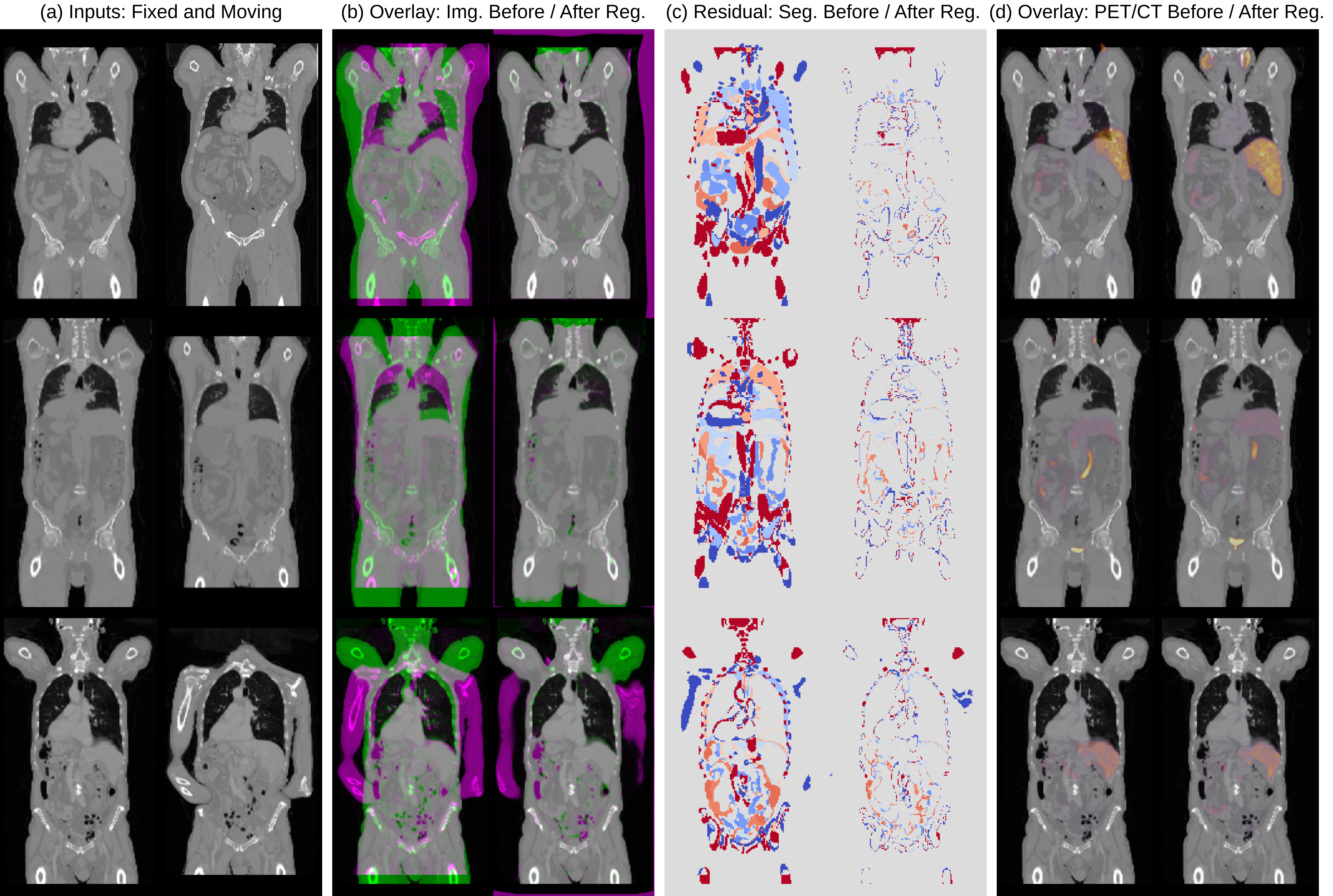}
    \caption{\textbf{Qualitative results}. On randomly selected input pairs (\textbf{a}), our framework increases visual, label, and cross-modal agreement (\textbf{b}, \textbf{c}, and \textbf{d}, respectively) across large deformations (rows 1 and 2), failing only in regions without topological correspondence (e.g., the arms in row 3).}
    \label{fig:qualresults}
\end{figure}

\section{Experiments}

\subpara{Dataset.}
We participated in the 2026 Learn2Reg~\cite{hering2022learn2reg,hansen2025learn2reg} PSMAReg~\cite{jeblick2026whole} Challenge. The pairs are longitudinal whole-body PSMA PET/CT scans that are downsampled by the challenge organizers to 192$\times$192$\times$288 at 2.7344$\times$2.7344$\times$3.27 mm resolution. The training set contains 202 image pairs, with TotalSegmentator annotations provided for the CT volumes and manual lesion annotations for the PET volumes. We do not train any models on this dataset; it is used only for hyperparameter optimization. 

\begin{table}[!t]
\centering
\caption{Performance of the top five methods on the test set of the challenge by rank, with scores, rankings, and evaluation provided by the challenge organizers. Rank confidence intervals are reported at the 95\% level. MTV and TLG errors are median percentage errors.}
\label{tab:test_summary}
\setlength{\tabcolsep}{4pt}
\resizebox{\linewidth}{!}{%
\begin{tabular}{lccccccccc}
\toprule
\textbf{Team} &
\textbf{Score} &
\textbf{Rank (95\% CI)} &
\textbf{CT Dice} &
\textbf{CT HD95} &
\textbf{PET Dice} &
\textbf{PET HD95} &
\textbf{MTV\%} &
\textbf{TLG\%} &
\textbf{NDV\%} \\
\midrule
Ours      & 71.4 & 1 (1--1)  & 0.830 & 4.6 & 0.610 & 19.4 & 1.78 & 1.39 & 0.0008 \\
MIAgent   & 37.3 & 2 (2--5)  & 0.832 & 4.9 & 0.628 & 18.9 & 5.37 & 5.35 & 0.0032 \\
tinymilky & 35.6 & 3 (2--11) & 0.832 & 4.6 & 0.598 & 20.0 & 3.41 & 3.24 & 0.0009 \\
shishuyue & 33.4 & 4 (2--6)  & 0.816 & 5.6 & 0.598 & 19.9 & 4.02 & 3.46 & 0.0273 \\
koegl     & 31.2 & 5 (3--9)  & 0.813 & 5.9 & 0.591 & 20.4 & 3.52 & 4.21 & 0.0180 \\
\bottomrule
\end{tabular}%
}
\vspace{-2em}
\end{table}

\subpara{Evaluation.} The challenge scores CT alignment by the DSC and HD95 of TotalSegmentator labels, and lesion alignment by the DSC and HD95 of manual PET lesion masks. It scores biomarker preservation by the MTV and TLG errors~\cite{larson1999tumor}, and regularity by \%NDV~\cite{liu2024finite}. The challenge ranks methods by a weighted geometric mean of Wilcoxon-based significance scores for registration accuracy (40\%), PET biomarker preservation (40\%) and field  regularity (20\%, with \%NDV $\leq$ 0.005\% counted as equal).

\subpara{Results.} Our framework ranks first on the PSMAReg challenge test set leaderboard (Table~\ref{tab:test_summary}). Sample alignments are illustrated for three randomly chosen subjects from the training set in Figure~\ref{fig:qualresults}. A detailed analysis is provided by the challenge organizers at \url{https://github.com/JHU-MedImage-Reg/Learn2Reg26_PSMA/tree/main/PSMAReg_TestPhase_Results}. %

\subpara{Ablations.} Table~\ref{tab:ablations_se} validates the main components of our framework on 65 ground-truth pairs from 50 held-out training subjects. Tuning FireANTs substantially improves registration over its default hyperparameters, while adding anatomix and MIND-SSC features as input provides a large gain in anatomical accuracy and PET biomarker preservation. In all cases, volume regularization and projection progressively reduce MTV/TLG error with little loss in alignment accuracy. In Stage 2, segmentation-guidance improves organ alignment (particularly in the abdomen), while adding anatomix supervision back in prevents the degradation of unsupervised tissue alignment (as measured by increase in CT MAE) observed when optimizing with labels alone. We also find that running the entire pipeline \uline{without} the anatomix and MIND-SSC features degrades performance.

\begin{table}[!t]
\centering
\caption{Ablation study on 65 training set pairs (50 held-out subjects) with mean and std. error.}
\label{tab:ablations_se}
\setlength{\tabcolsep}{4pt}
\resizebox{\textwidth}{!}{%
\begin{tabular}{@{}l c c c c c c@{}}
\toprule
\textbf{Configuration} & \textbf{CT DSC} $\uparrow$ & \textbf{CT HD95} $\downarrow$ & \textbf{\%NDV} $\downarrow$ & $|$\textbf{MTV\%}$|$ $\downarrow$ & $|$\textbf{TLG\%}$|$ $\downarrow$ & \textbf{CT MAE} $\downarrow$ \\
\midrule
No registration             & 0.114(0.011) & 41.72(2.98) & -- & -- & -- & 302.1(12.0) \\
Stage 1, CT+PET, default FireANTs     & 0.586(0.027) & 20.07(3.43) & 0.00025(0.00015) & 15.49(2.15) & 16.67(2.31) & 118.1(14.5) \\
Stage 1, CT+PET, tuned FireANTs & 0.686(0.012) & 11.09(0.94) & 0.00113(0.00086) & 15.56(2.08) & 18.47(2.39) & \textbf{57.3}(3.3) \\
\quad + anatomix + MIND-SSC & 0.775(0.007) & 6.54(0.37) & 0.00125(0.00085) & 6.31(1.68) & 6.93(1.76) & 58.6(3.8) \\
\quad + volume penalty  & 0.773(0.007) & 6.57(0.36) & 0.00145(0.00090) & 4.95(1.28) & 5.36(1.48) & 60.0(3.8) \\
\quad + volume projection      & 0.772(0.007) & 6.57(0.36) & 0.00147(0.00092) & 3.41(1.18) & 4.14(1.49) & 60.5(3.8) \\
Stage 2, labels only (no image term)   & 0.808(0.005) & 5.40(0.28) & 0.00304(0.00094) & 5.09(1.07) & 5.16(1.32) & 78.3(3.5) \\
\quad + anatomix & 0.809(0.005) & \textbf{5.39}(0.28) & 0.00288(0.00094) & 5.64(1.23) & 5.88(1.60) & 62.0(3.2) \\
\quad + volume projection + smoothing & \textbf{0.812}(0.006) & 5.45(0.29) & \textbf{0.00014}(0.00006) & \textbf{2.99}(0.96) & \textbf{3.85}(1.44) & 62.0(3.2) \\
\midrule
Full pipeline w/o anatomix+MIND-SSC & 0.797(0.007) & 6.61(0.48) & 0.00020(0.00008) & 3.60(1.43) & 4.57(2.18) & 58.8(2.5) \\
\bottomrule
\end{tabular}%
}
\vspace{-1em}
\end{table}

\section{Discussion}

\subpara{Limitations and future work.} We use a single motion model to represent several disparate ones across the body. As a result, this model is underconstrained when fitting articulated motion (e.g., of the limbs) across time points and overconstrained when there is no topological correspondence. For example, when the arms are crossed in the baseline and are above the head and outside the field-of-view in the follow-up, diffeomorphic warps cannot fit this motion without topology changes~\cite{niethammer2011geometric,trouve2005metamorphoses}. Future work could use bone-specific~\cite{gopalakrishnan2026polypose,yip2014development} polyrigid models~\cite{arsigny2005polyrigid} to fit limb articulation.

\subpara{Conclusions.} 
Generalist models trained on large aggregated collections already exist for medical image registration~\cite{tian2024unigradicon}. However, to our knowledge, they generally have not yet exceeded the performance of custom models trained on the specific tasks. We instead jointly leverage the inductive biases of high-performing registration solvers with high-dimensional representations of images extracted by foundation models pretrained entirely on synthetic anatomy, and achieve the best test-set result despite never having trained on the challenge's training set. We believe that hybrid approaches such as ours that combine networks pretrained on large collections of wildly variable synthetic data with high-quality solvers provide a scalable path for practitioners to align their datasets without needing to train networks for each new task.

\subsection*{Acknowledgments}

This work was supported by NIH grants R01 EB033773 and S10 OD038222 and an MGB Neuroscience Institute Transformative Scholar Award. We thank Vivek Gopalakrishnan, Jingru Fu, and Margherita Firenze for discussions and logistical help.

%
%
%
\bibliographystyle{splncs04}
\bibliography{refs}

\clearpage
\newpage

\section*{Appendix}

\subpara{Agent usage.}
We used AI agents (Claude Fable 5.0 and GPT 5.6 Sol) extensively for software engineering, hyperparameter tuning, and PET/PSMA-specific literature search. For example, they implemented the throughput and utilization improvements enabling the package container to meet the time and resource constraints defined by the challenge. Further, the volume projection method used here~\cite{hameeteman2008volume} was found and implemented on the GPU by prompting them to search for and implement relevant published methods for post-processing based volume preservation in registration. We also used them to compile the list of organs with high physiological PSMA uptake when constructing the high-recall lesion mask.  
As of August 2026, we have not yet found them suitable for suggesting specific technical directions in the registration methods themselves without considerable steering. 

\subpara{Implementation details.} 
We clip the CT to [$-1024$, 1024]~HU and the PET to [0, 15]~SUV, and min-max normalize each. The body mask is computed without a network: we threshold the CT at $-500$~HU, erode by one voxel to separate the patient from the scanner table, keep the largest connected component, fill holes slice-wise, and apply a two-voxel morphological closing. This takes 0.7~s and agrees with TotalSegmentator's body model at a Dice of 0.993--0.995. anatomix runs on $128^3$ windows with 25\% overlap in FP16, skipping windows that contain no body. Stage 1 uses at most 100, 100, 100, and 30 Adam iterations per level, and each level stops early when the loss no longer decreases. LION~\cite{Piresjnumed.126.272852} uses 6 of its 8 mirrored test-time augmentation passes. 

For the Stage-2 organ labels, for speed, we run TotalSegmentator's~\cite{wasserthal2023totalsegmentator} 3-mm model concurrently with two 1.5-mm models restricted to tight crops, one for the 52 rib and vertebra classes and one for the remaining 65 classes, and merge their outputs into a single 117-class map; all models use a tile step of 0.8. The lesion mask uses the 6-mm LION model. All local computation was performed on NVIDIA RTX PRO 6000 Blackwell GPUs with 96~GB of memory, using CUDA 13.0 and PyTorch 2.13.0+cu130.

\subpara{Organs with physiological PSMA uptake.} We subtract the following TotalSegmentator classes from the SUV-thresholded mask: \texttt{brain}, \texttt{heart}, \texttt{esophagus}, \texttt{stomach}, \texttt{duodenum}, \texttt{small\_bowel}, \texttt{colon}, \texttt{liver}, \texttt{gallbladder}, \texttt{pancreas}, \texttt{spleen}, \texttt{kidney\_left}, \texttt{kidney\_right}, \texttt{kidney\_cyst\_left}, \texttt{kidney\_cyst\_right}, and \texttt{urinary\_bladder}. Salivary and lacrimal glands also show high physiological uptake, but are not segmented by TotalSegmentator and are therefore not subtracted.

\end{document}